\documentclass[letterpaper, 10 pt, conference]{ieeeconf}  
\makeatletter
\def\endthebibliography{%
  \def\@noitemerr{\@latex@warning{Empty `thebibliography' environment}}%
  \endlist
}
\makeatother

\IEEEoverridecommandlockouts                              
\title{\LARGE \bf
Co-Design of Assistive Robotics with Additive Manufacturing and Cyber-Physical Modularity to Improve Trust}

\author{Alexandre Colle$^{1}$, Ronnie Smith$^{1}$, Scott MacLeod$^{2}$ and Mauro Dragone$^{2}$

\thanks{This work was supported in part by the Engineering and Physical Sciences Research Council (EPSRC) Centre for Doctoral Training (CDT) in Robotics and Autonomous Systems (grants EP/L016834/1 and EP/S023208/1) at Heriot-Watt University and The University of Edinburgh.}

\thanks{$^{1}$Alexandre Colle and Ronnie Smith are with the Edinburgh Centre for Robotics at Heriot-Watt University and The University of Edinburgh, Edinburgh UK. Emails:
        {\tt\small a.colle@sms.ed.ac.uk, ronnie.smith@ed.ac.uk}}%

\thanks{$^{2}$Scott MacLeod and Mauro Dragone are with Heriot-Wartt University and affiliated with the Edinburgh Centre for Robotics, Edinburgh, UK. Emails:
        {\tt\small sam19@hw.ac.uk, m.dragone@hw.ac.uk}}%
}

\usepackage{cite}
\usepackage{graphicx}

\begin{document}

\maketitle
\thispagestyle{empty}
\pagestyle{empty}

\begin{abstract}
Robotics and automation have the potential to significantly improve quality of life for people with assistive needs and their carers. Adoption of such technologies at this point in time is far from widespread. This paper presents a novel approach to the design of highly customisable robotic concepts, embracing modularity and a co-design process to increase the involvement of end-users in the development life cycle. We discuss this process within the context of an elderly care use case. Using design methodology and additive manufacturing, we outline how key stakeholders can be involved from initial conception through to integration of the final product within their environments. In future work, we will apply this process to demonstrate the effectiveness of our approach for improving long-term acceptance and trust of robotic technology in care contexts.
\end{abstract}

\section{Introduction}

Health and elderly care systems in many developed countries have been facing significant pressure for several years \cite{Donnison2018}: budget cuts, a lack of human resources, and a growing elderly demographic have tested the boundaries of existing care systems \cite{Vaughn2019}. The recent COVID-19 crisis has increased this pressure further, putting the lives of both care workers and care recipients, at risk.

We have seen robotics applied to various aspects of care, such as in minimally-invasive surgery and in automated cleaning of facilities \cite{Pang2018}. The integration of robotics into existing care practice nonetheless remains difficult for several reasons, which we explore herein.

Providing high-quality care, whether for healthcare in general or for elderly care in particular, requires significant human effort. The demands and requirements of care staff may differ greatly from those of patients and residents \cite{Vaughn2019}. Furthermore, the introduction of new technology can be difficult to accept, which can lead to misuse of or outright refusal to interact with Assistive Technology (AT).

Current robotic platforms, in the context of care, tend to provide very specific solutions to users. These robots represent completed products, with development teams often removed from the reality of users. These devices are subject to manufacturing processes and commercialisation constraints, often in conflict with systemic problems of stakeholders in the context of care \cite{Maibaum2021}. The initial manufacturing costs are often substantial.

Understanding the needs and wants of stakeholders is fundamental to the development of meaningful AT. User-Centred Design (UCD) and `Co-Design' can improve the design of assistive technology for elderly care, by placing the users at the core of the design process. At the same time, co-design processes encourage human interactions and social contact, thus increasing trust between researchers and participants \cite{Schwaninger2021}\cite{Duque2019}. However, researchers need to adapt to the specific requirements of older users with specific tools appropriate for an audience less well-versed in technology. 
Co-design has been adopted largely in recent years in the field of Human-Robot Interaction (HRI)  \cite{Randall2018}\cite{Winkle2020}\cite{Lee2017a}\cite{Tonkin2018}, however, few projects have led to the design of tangible robots with a practical use case. Service robots remain very complex technological systems. Their development requires multidisciplinary teams, working closely to achieve a common goal with minimum resources.

Research in Artificial Intelligence (AI) has led to improved cognitive capacity in robots. In contrast, as their minds evolve, their physical form remains largely unaltered \cite{Bongard2014}. The current way robots are designed is prohibitive of physical alteration and discourages active personalisation. Customisation, the process of letting users actively modify the robot appearance, includes users further in the development of the robot in the long-term. When users are involved with customisation, it increases customer satisfaction, integration, and loyalty \cite{THEILMANN2014}. For robotics, this process can have clear benefits for social interactions \cite{Dragone2015}, and encourage active participation and use of a co-created, customisable, device.

According to Gibson et al. \cite{Gibson2019}, in the context of dementia care, users tends to ``tame'' technological artefacts by adapting them to their individual needs. Gibson et al. refer to this practice as \textit{bricolage}: where carers use creative and non-conventional ways to use AT to adapt to local needs. They also highlight the frustration of these same carers, where their specific input does not materialise in final products. As such, bricolage highlights the importance of constant design iterations in the context of care, in order to create devices matching the vast array of needs from both carers and those receiving care. It also emphasises the risk of developing narrow technological solutions in conflict with the reality of users requirements. In contrast, the concept of bricolage encourages provider of AT solutions to facilitate user-driven customisation, instead of current standardised solutions \cite{Gibson2019}.  Above all, it validates the need for a systemic method such as co-design, to capture this iterative process, and translate these suggestions into tangible outcomes. 

We hypothesise that the outcomes of co-design can be improved through the use of Additive Manufacturing (AM) and a modular approach to robot development, ultimately improving acceptance and trust in the robot.

Our long-term vision is to develop a framework to guide the development of a robot combining co-design and AM. This paper explores the role of co-design and customisation as a mediator to improve the long-term acceptance of robots in a care setting. We also discuss the role of AM in improving the conception of modular robots to facilitate the participatory design work.

\textit{Research Question:} how can we more effectively design these new types of robots?

\section{Benefits of Modular Robots}

Current commercial robots have been effectively adopted in various environments. The acquisition of these machines represents a major investment due to their complexity, however, their flexibility at executing tasks has great advantages. A new generation of service robots is challenging the fundamental nature of these machines. Contrary to traditional robots, modular robots offer a better solution to unpredictable and evolving tasks \cite{Ahmadzadeh2016}. For example, a stackable, re-configurable and adaptable concept for monitoring oil-rigs named Limpet II \cite{Sayed2020}, allows users to swap modules to execute different tasks. Instead of a uniform and expensive solution provided by traditional robots, the Limpet II creates a low cost and adaptable solution to achieve multiple tasks within a complex environment \cite{Sayed2020}.  Modular robots raise the prospect of fully adaptable devices, in constant evolution with users needs. 

\subsection{Improving Flexibility}

The development process behind individual robots implies strong financial capabilities from robotics companies \cite{Schmelzer2019}. Research and Development (R\&D) represents a substantial cost, and market readiness absorbs a large part of the initial resources. Additionally, a robotics company depends on the capacity and willingness of a manufacturing partner to produce final products within a specific price range. Thus, Social robots, in the context of healthcare, suffer from a lack of flexibility in product development because the manufacturing process depends on traditional manufacturing methods.

Progress in AM techniques and the improvement of Computer-Aided Design (CAD) software capabilities are transforming manufacturing. AM, is ``the processes that add some materials to the previous surface
via different deposition techniques that lead to different part quality, density, and geometrical accuracy'' \cite{Mehrpouya2019}. This `Industry 4.0' \cite{Mehrpouya2019} is faster, more sustainable \cite{Niaki2019}, and more efficient \cite{Attaran2017}. CAD allows for greater team collaboration regardless of individual locations, while managing and optimising extremely complex designs \cite{Nordin2018}. In conjunction with CAD, AM is breaking the chain of traditional manufacturing. As a result, three-dimensional (3D) printing supports rapid prototyping and local manufacturing of those designs \cite{Ziokowski2020}.

AM and CAD create new channels to develop robots with flexibility as a principle of development. As AM technology improves, the development of flexible robotics can reap the benefits by applying the advancements to areas such as rapid prototyping and manufacturing. More generally, modular robotics is getting more attention from research and the private sector for their advantages in terms of ``reconfigurability, reusability, and ease in manufacturing'' \cite{Chennareddy2017}\cite{Sayed2020}. Modularity can take many shapes and forms and depends on the specific area of use of the robot. Ultimately, a clear understanding of the context of use will inform how modularity should be applied.  

\subsection{Improving Acceptance}

Maintaining acceptance levels of any given robot in the long-term, particularly in a public environment, remains a challenge for HRI. One of the goals of HRI is to improve the long-term acceptance of social robots \cite{Dragone2015}\cite{Glende2016}\cite{Turja2020}\cite{Peek2017}\cite{Naneva2020}. Robots as an object are perceived by different individuals in different ways. From patients to health workers, robots can either be seen as a threat or an opportunity. The way robots are developed at the moment does not encourage active public participation in the eventual behaviours and overall appearance of the robot.

Of course, robots are not positively perceived by everyone. Individual taste varies from one person to another, as aesthetic judgement is made out of both conscious and subconscious decisions \cite{Haug2016}. Accordingly, the opinion of an audience will vary greatly when confronted with robots with a uniform design. Moreover, a moving agent will increase anthropomorphic interpretation and the impression of agency \cite{Waytz2010d}. Depending on the context, these interactions can have overwhelmingly positive or negative outcomes such as fear and total rejection \cite{Yogeeswaran2016}. 

Technology acceptance differs between age groups and the integration of complex technology varies with the context of use. For an end-user, taking an active part in the design and functionality of a complex device like a robot improves the chances of acceptance and consistent long-term use \cite{Peine2011}\cite{Sung2009}. Moreover, giving users the ability to engage in personalisation encourages them to adapt robots to their specific circumstances \cite{Peine2011}, improving relationships with the device at the same time. Peine et al. \cite{Peine2011} insist technology must be domesticated or ``tamed'' and that ``Users are more likely to become co-designers of technology to fit complex technical arrangements into their lives''. The motivations behind `bricolage' and personalisation need to be captured, analysed and transcribed into concrete outcomes, for both the benefit of users and developers of the robots. 

Current robotic platforms are not designed to let users actively personalise their appearance and function. Modularity must be considered at the conception of the device, its implication in terms of hardware and software are very complex \cite{Chennareddy2017} and cannot be treated as a cosmetic add-on. 

\subsection{Co-Design}

The process of design thinking, as a support tool, allows for a holistic approach to problem-solving and product development \cite{Brenner2016}\cite{Tschimmel2012}\cite{PlattnerH.LeiferL.Meinel2011}. co-design, naturally involves stakeholders and  \cite{Scariot2012}\cite{Gulliksen2003} allows for great participation of the users involves. 

Co-design is executed in several stages, consisting of generally these main steps \cite{Blessing2009a}:
\begin{enumerate}
    \item Empathise with users;
    \item Define problems;
    \item Ideate (idea generation);
    \item Prototype concept;
    \item Test and iterate.
\end{enumerate}

Applied in the context of robotics, Compagna et al. used a participatory approach for the development of a service robot in \cite{Compagna2015} elderly care. Their three years study use the stages of co-design to inform the development of their robot. The results ultimately recommended a substantial increase in usability trials combined with better use of rapid prototyping. Additionally, they recommended to encourage direct relations between developers and stakeholders and avoid intermediary actors in the process.


Initially, the thought of robots may disturb the public ability to imagine rationally what could be an autonomous device\cite{Herrmann2013}\cite{Banks2020} in their environments. The influence of science fiction raises people's expectations of the real capability of machines. Moreover, many people experience feelings of fear and aversion to technology when confronted with the thoughts of robots\cite{Stein2019}. The label `\textit{robot}' can be misinterpreted, therefore there are some cases where a different denomination \cite{Kim2020} will ease people’s imagination.

Additionally, broader concerns can create additional frictions. For stakeholders in the context of ageing, being seen as a sick patient in need of care will raise suspicions and have a negative impact on a co-design initiative \cite{Stein2019}. On the contrary, if they become active participants working for their own benefits and potentially helping others, they will then appropriate the project for themselves and become far deeper involved as participants \cite{Peine2011}. The elderly want to be treated like normal citizens, and considered fairly, with their own needs \cite{Kohlbacher2011}\cite{Peine2011}. By contrast, carers will have personal concerns against automation. They worry about the future of their career \cite{Teo2020} and towards the people they care about. They believe more automation will lead to more unemployment and patients being neglected. Stakeholders across the board must be understood and we must adapt our approach to anticipate negative feelings. The goal of this project is to allow them to create their tools to improve their work and well-being and at the same time improve trust in robots.

\section{A Modular Robotics Platform for Co-Design}

\begin{figure}[ht]
\centering
\label{graph}
\includegraphics[width=\linewidth]{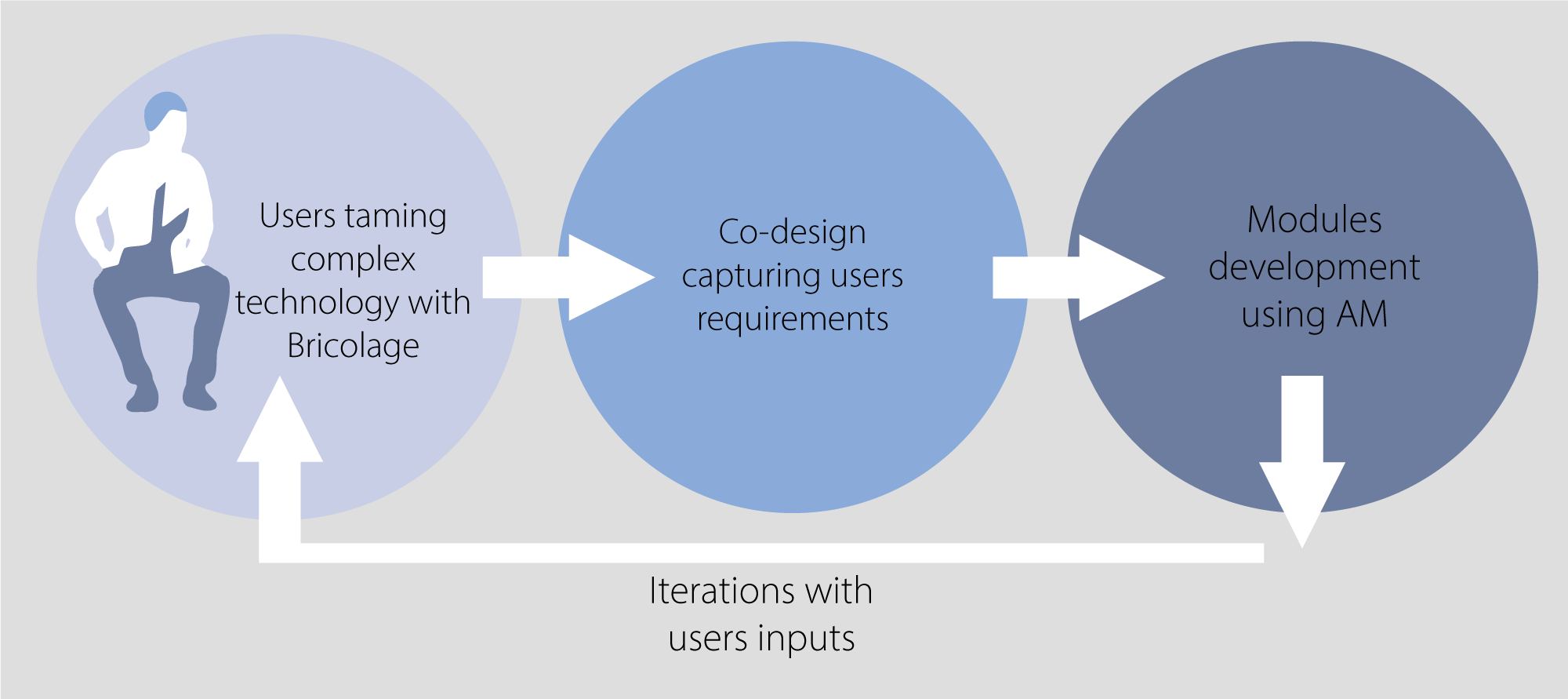}
\caption{Integration of co-design methods and Additive Manufacturing development strategy.}
\end{figure}

Having reviewed the benefits and best-practice of co-design, we now lay out our concept of pairing it with a modular robotic platform and additive manufacturing to enable rapid development and manufacture of co-designed modules. We posit that this will enable significantly tighter iterative cycles of feedback and development, therefore enabling sustained user engagement in the process as the products of their feedback is more quickly realised (see Figure \ref{graph}).

We envisage that such a platform to enhance this process would consist of a `base' robot with minimal functionality, but to which modules that provide additional functionality can be readily attached. There is therefore a fine balance to be struck between the functionality provided by the base and the range of functionality that can be provided through modules.

In this scenario, the platform will become a combination of top-down (prescribed) and bottom-up (learned, through co-design) design \cite{yew_trust_2020}. The balance between the two types of design is of the utmost importance, but there should be a bias towards the bottom-up approach. As such, a requirement for this modular platform is that the base should provide minimal functionality, the essential traits to be considered a `robot', and can incorporate additional hardware that might be useful (e.g. computational resources, cameras, speakers) for use by modules, without being prescriptive about what that use is.


In terms of hardware and software, this means the platform must be flexible such that a newly developed module can be easily attached the to the base and the behaviour and functionality of the module be realised in software. This should draw from concepts in Component-Based Software Engineering (CBSE) in that new features can be gained through `plug and play' of new modules.

In contrast with traditional platforms, our method can produce, in a limited time-frame, a realistic and deployable robotic system within users' environments. The modular nature of the device is naturally designed for continuous personalisation.

We are suggesting to emphasis the following steps for the development of this device:

\begin{itemize}
     \item \textbf{Method:} Methods such as design thinking help structuring a project around the same goal. It also represents a similar language when multidisciplinary teams collaborate, by providing a central canvas for stakeholders and developers can engage in a ``design Conversation'' \cite{Brenner2016a};
     \item \textbf{Conception:} Advancements in additive manufacturing technologies improves not only rapid prototyping as part of the co-design method, it also allows for the rapid manufacturing of a finished product \cite{Attaran2017a}. Our goal is to use the flexibility offered by additive manufacturing to rapidly execute the changes highlighted during the co-design sessions. The results can offer a tangible result and create a concrete outcome for participants. Ideas become real, therefore increasing trust between organisation and participants \cite{Blomkvist2014};
     \item \textbf{Design:} The appearance and behaviours of a social robot have a major impact on how negatively or positively a robot will be perceived by users \cite{Hegel2012}. Therefore, the design of the appearance of the robot is a crucial part of the design process. By contrast, this robot represents a new design challenge and transgresses the current paradigm of design. As a result, we advocate a new design principle, where we purposefully limit the final appearance of the robot. Following our hypothesis, final users will then be encouraged to finish the overall design of the robot by engaging in active personalisation. For this purpose, we employ an open-ended design methodology, leaving the final outcome forever open, never completely finished \cite{Dubberly2019}. As mentioned earlier, the shape of robots needs to evolve in relation to their cognitive capacity, and also with the needs of the users. Both have unlimited room to grow. 
\end{itemize}

\section{Conclusion \& Future Work}

In this paper, we have highlighted the pitfalls of current assistive robotic solutions, particularly for use with the elderly. We have also highlighted the potential benefits of flexible manufacturing techniques, such as Additive Manufacturing, and emphasised the promise of the co-design process.

On that basis, we have made recommendations about how the outcomes of co-design of assistive robotics can be improved through iterative cycles of idea generation, rapid development, and feedback. Furthermore, we have highlighted how this may improve overall acceptance and trust in robotics.

In our future work, we will put our suggested process into practice as we develop a modular robotic platform that will serve as a sandbox for the co-design of useful assistive modules with our partners in the health and social care industries.


The co-design process will begin with some fieldwork and observation on-site. Co-design implies collaboration and a long-term engagement from the researchers and the stakeholders. As such, we plan to start the process by using semi-structured interviews to engages with health and social care stakeholders, involving carers and care and support service users, in a direct and open conversation. This process personifies the research, offers the opportunity of social bond and facilitates long-term collaboration.



\bibliographystyle{IEEEtran}
\bibliography{ronniezotero,alex_library}

\end{document}